\UseRawInputEncoding
\documentclass[letterpaper, 10 pt, journal, twoside]{IEEEtran}

\usepackage{graphics} 
\usepackage{float}
\usepackage{subfigure}
\usepackage{epsfig}
\usepackage{mathptmx}
\usepackage{caption}
\usepackage{times}
\usepackage{amsmath} 
\usepackage{amssymb}  
\usepackage{balance}
\usepackage{booktabs}
\usepackage{authblk}
\usepackage{soul,color}
\usepackage{multirow}
\usepackage{ulem}
\usepackage{hyperref}
\usepackage{cite}
\usepackage{lettrine}
\usepackage[moderate]{savetrees}

\makeatletter
\def\@citex[#1]#2{\leavevmode
\let\@citea\@empty
\@cite{\@for\@citeb:=#2\do
{\@citea\def\@citea{,\penalty\@m\ }%
\edef\@citeb{\expandafter\@firstofone\@citeb\@empty}%
\if@filesw\immediate\write\@auxout{\string\citation{\@citeb}}\fi
\@ifundefined{b@\@citeb}{\hbox{\reset@font\bfseries ?}%
\G@refundefinedtrue
\@latex@warning
{Citation `\@citeb' on page \thepage \space undefined}}%
{\@cite@ofmt{\csname b@\@citeb\endcsname}}}}{#1}}
\makeatother

\title{Visual-Tactile Cross-Modal Data Generation using\protect\\Residue-Fusion GAN with Feature-Matching\protect\\and Perceptual Losses}

\author{Shaoyu Cai$^{1}$,   
        Kening Zhu$^{1,2,*}$,
        Yuki Ban$^{3}$, 
        Takuji Narumi$^{4}$

\thanks{Manuscript received: February, 24, 2021; Revised June, 5, 2021; Accepted July, 1, 2021. 

This paper was recommended for publication by Editor Jee-Hwan Ryu upon evaluation of the Associate Editor and Reviewers' comments.

This research was partially supported by the Young Scientists Scheme of the National Natural Science Foundation of China (Project No. 61907037), the Guangdong Basic and Applied Basic Research Foundation (Project No. 2021A1515011893), the Applied Research Grant (Project No. 9667189), and ACIM, School of Creative Media, City University of Hong Kong. This research was also partially supported by JSPS KAKENHI Grant (Number 20K21801 and Number 21H03478).} 

\thanks{$^{1}$School of Creative Media, City university of Hong Kong, Hong Kong, P. R. China {\tt\small shaoyu.cai@my.cityu.edu.hk}}%
\thanks{$^{2}$City University of Hong Kong Shenzhen Research Institute, Shenzhen, P. R. China {\tt\small keninzhu@cityu.edu.hk}}
\thanks{$^{3}$Graduate School of Frontier Sciences, The University of Tokyo, Chiba, Japan
        {\tt\small ban@edu.k.u-tokyo.ac.jp}}
\thanks{$^{4}$Graduate School of Information Science and Technology, The University of Tokyo, and JST PRESTO, Tokyo, Japan 
        {\tt\small narumi@cyber.t.u-tokyo.ac.jp}}%

\thanks{$^{*}$Corresponding author: Kening Zhu}%

\thanks{Digital Object Identifier (DOI): see top of this page.}
}

\begin{document}
\maketitle

\begin{abstract}

Existing psychophysical studies have revealed that the cross-modal visual-tactile perception is common for humans performing daily activities. However, it is still challenging to build the algorithmic mapping from one modality space to another, namely the cross-modal visual-tactile data translation/generation, which could be potentially important for robotic operation. In this paper, we propose a deep-learning-based approach for cross-modal visual-tactile data generation by leveraging the framework of the generative adversarial networks (GANs). Our approach takes the visual image of a material surface as the visual data, and the accelerometer signal induced by the pen-sliding movement on the surface as the tactile data. We adopt the conditional-GAN (cGAN) structure together with the residue-fusion (RF) module, and train the model with the additional feature-matching (FM) and perceptual losses to achieve the cross-modal data generation. The experimental results show that the inclusion of the RF module, and the FM and the perceptual losses significantly improves cross-modal data generation performance in terms of the classification accuracy upon the generated data and the visual similarity between the ground-truth and the generated data.

\end{abstract}

\IEEEpeerreviewmaketitle
\section{Introduction}
\lettrine[lines=2]{V}{ision} and touch are two important sensory channels for humans perceiving and understanding the world \cite{yau2009analogous}. Through vision, we can observe the environment and understand the appearances of objects, such as surface patterns, sizes, shapes, and colours. Besides, we can directly interact with the object surface through touch, and perceive the surface material and texture of an object. However, research shows that it is challenging for humans to gain a thorough understanding of an object through the only single sensory channel (either vision or touch) \cite{guest2003role}. While lacking the information from a certain perceptive modality, humans usually need to and can perform cross-modal perception estimation. That is, imagining the feeling of one perceptional channel according to the real sensation from another channel. For instance, while seeing the image of a textured surface (e.g., glossy and rough), we can estimate how it feels like for touching (e.g., smoothness). We can also imagine the appearance of a textured surface without seeing it while touching or sliding our fingers on it. Such cross-modal perception connects the visual and haptic sensations, and can improve the capability of scene/object recognition for humans \cite{calvert2001crossmodal, newell2005visual}. 

\begin{figure}[t]
\centering
\subfigure[Tactile-to-Visual (T2V) Data Generation]{
\begin{minipage}{\linewidth}
\centering
\includegraphics[width=\columnwidth]{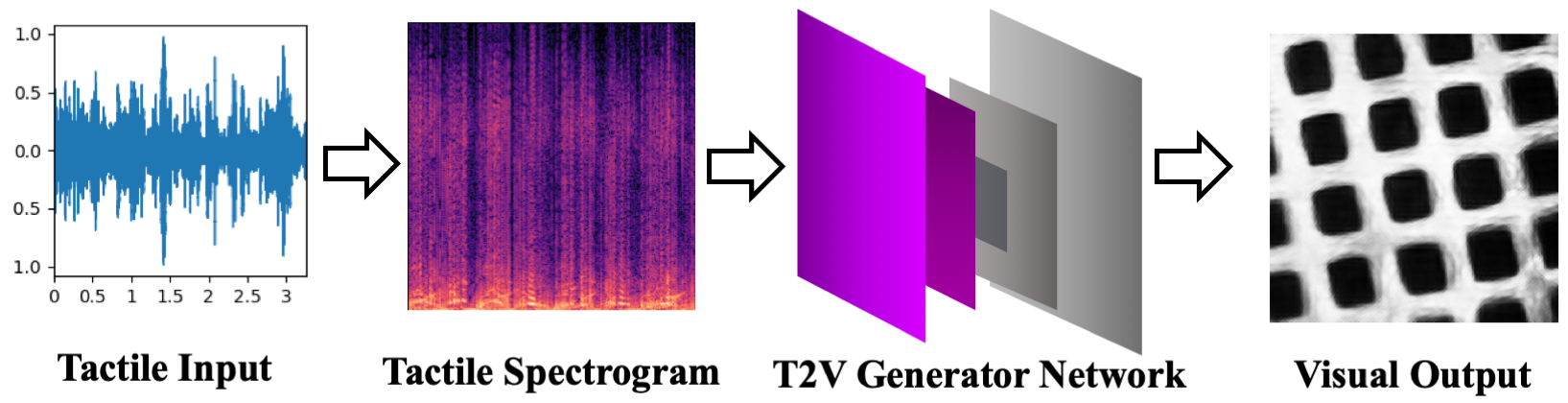}
\vspace{-0.15in} 
\end{minipage}%
}%
\vspace{-0.1in} 
\subfigure[Visual-to-Tactile (V2T) Data Generation]{
\begin{minipage}{\linewidth}
\centering
\includegraphics[width=\columnwidth]{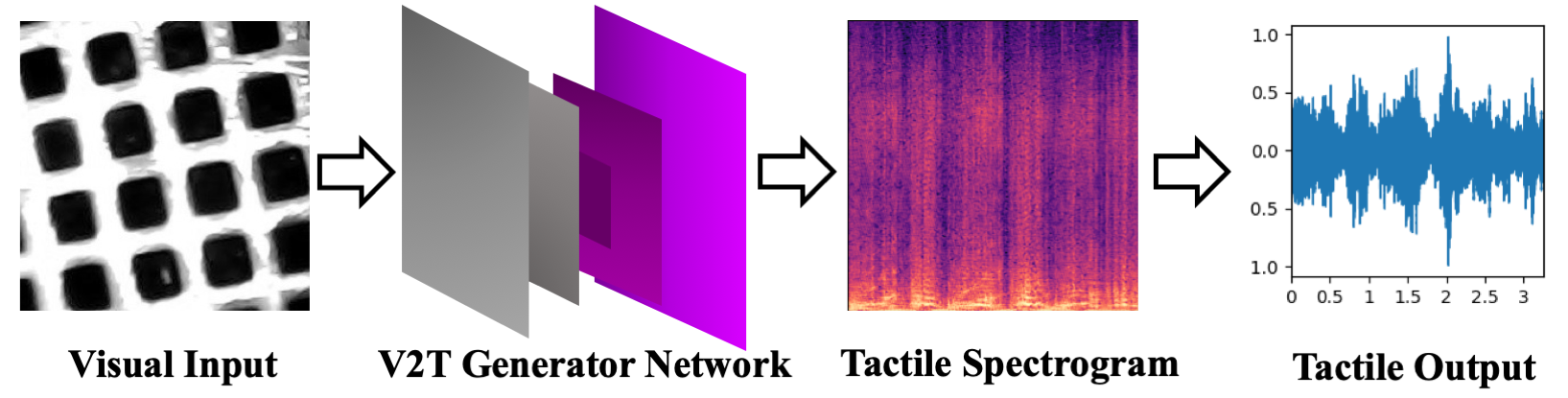}
\vspace{-0.15in} 
\end{minipage}%
}%
\vspace{-0.05in} 
\caption{The workflow of (a) T2V: Tactile-to-Visual and (b) V2T: Visual-to-Tactile cross-modal data generation. Here we use the grey-scale surface-texture images for the \textbf{visual domain} and the amplitude spectrograms of the time-series acceleration signals from the on-surface pen-sliding movement for the \textbf{tactile domain}. For the tactile data, the time-series signal could be converted to and generated from the spectrogram using the algorithm of Short-Time Fourier Transform (STFT) and the Griffin-Lim algorithm \cite{griffin1984signal} respectively.} 
\vspace{-0.1in}
\label{Fig.1} 
\end{figure}

For robotic operation, vision-based sensing technology has been widely applied for various tasks, such as object detection \cite{karaoguz2019object}, object tracking \cite{markovic2014moving}, object grasping \cite{fang2018dual}, and navigation \cite{wellhausen2020safe}. Additionally, robots with haptic sensors (e.g., accelerometer, gyroscope, thermochromic-based tactile sensor \cite{sun2019novel}, GelSight sensor \cite{yuan2017gelsight}, etc.) could perform the touch-related tasks, such as texture recognition \cite{sinapov2011vibrotactile} and grasping objects in different shapes \cite{calandra2017feeling} and hardness \cite{yuan2017shape}. As the vision and the tactile modalities often provide complementary information to each other, recent works argue that the usage of a single sensory modality may limit the operational capabilities of the robots in unstructured environments \cite{tatiya2019deep, bajcsy2018revisiting}. To mimic human's capability of cross-modal perception, there have been recent works focusing on the integration of and the conversion between the visual and the tactile data, such as generating the spectrograms of the vibration signals from the visual surface images \cite{ujitoko2018vibrotactile, 9018269}, and generating the GelSight-based image according to the visual image of an object surface and vice versa \cite{lee2019touching, li2019connecting}. Such visual-tactile data generation could be potentially applied to provide haptic feedback and enable the gestural interface in human-robot interaction and remote communication \cite{barber2013visual, che2018facilitating}.

To further explore the integrated visual-tactile perception for robotics, we present a deep-learning-based framework for the cross-modal visual-tactile data generation (Fig. \ref{Fig.1}). The presented framework is built upon the base of the generative adversarial networks (GANs) with the residue-fusion (RF) module, and trained with the additional feature-matching (FM) and perceptual losses. Compared to the existing works on GAN-based visual-tactile data generation between the GelSight and the visual images \cite{lee2019touching, li2019connecting}, we focus on the vibrotactile signal of the accelerometer, which is considered to be lower-cost and also widely used for robotic texture recognition \cite{sinapov2011vibrotactile, jamali2011majority}. More importantly, the accelerometer-based vibrotactile signal shows a more significant difference in the spatial and the temporal domains towards the visual image data than the GelSight image-based tactile data does. Therefore, the existing solutions of image-to-image translation may not be directly applicable. Our generative framework is trained upon the visual and the tactile data of 9 types of materials selected from the LMT-108 database \cite{7737070}. Our experiments show that the proposed framework could generate the visual and the tactile data that are visually and statistically more closed to the ground truth than the baseline Pix2Pix model \cite{isola2017image} that has been used for cross-modal data generation \cite{lee2019touching}. In terms of the algorithmic/robotic perception, the generated data could be classified by the pre-trained visual- and tactile-signal classifiers with considerable accuracies (visual data: 94.61\%; tactile data: 83.78\%). Our source code and data set are available at: \url{https://github.com/shaoyuca/Visual-Tactile-Data-Generation}.

\section{Related Works}
Our work is highly inspired by the existing works on cross-modal learning, specifically visual-tactile data generation.

\subsection{Cross-Modal Learning}
Our work falls under the umbrella of cross-modal learning. Cross-modal learning usually extracts the shared information and construct the association between two different modality domains. For instance, SoundNet \cite{aytar2016soundnet} directly processes the audio data in waveform and optimizes the Kullback–Leibler distance of feature representations between the video and the audio. Owens et al. \cite{owens2016visually} propose the recurrent-neural-network-based (RNN-based) algorithm for sound prediction from silent videos. Liu et al. \cite{liu2020audiovisual} present an audio-visual cross-modal retrieval system to retrieve materials across the visual and the auditory data. Yuan et al. \cite{yuan2017connecting} associate the colour and the GelSight-based tactile images of the fabric samples by jointly training two convolutional neural networks (CNNs) across these two types of data. To this end, Yuan et al. show that cross-modal learning with the jointly trained subspace could obtain the shared features in two different modalities, indicating the feasibility of cross-modal visual-tactile data generation.

Recently, Generative Adversarial Networks (GANs) \cite{goodfellow2014generative} show the capability of cross-modal data generation. Within the visual domain, the Pix2Pix \cite{isola2017image} and the CycleGAN \cite{zhu2017unpaired} models have been widely used for image-to-image translation, such as sketch-based photo generation \cite{chen2020deepfacedrawing}. As a signal can be represented in the image/matrix format in the time and the frequency domains, researchers have experimented with the potential of these image-generation models on constructing the mapping between two different modalities, such as sounds and images \cite{duan2019cascade}, images and touch \cite{ve.20201254}, and sounds and videos \cite{wan2019towards}. Adopting the conditional GAN structure, Chen et al. \cite{chen2017deep} present a two-way generation framework for audio-to-visual and visual-to-audio generation. Generally, image-to-image translation assumes the geometrical alignments between inputs and outputs, which shows poor results on domain adaptations with significantly scale difference between two domains, such as the visual and the tactile domains \cite{li2019connecting}. Cross-modal data retrieval could be one alternative approach, besides the GAN-based cross-modal data generation. Using cross-modal data retrieval, for example, taking one spectrogram-based audio signal as input, the trained feature extractors could retrieve a visual surface image with the most matching features in the training database, and vice versa \cite{liu2020audiovisual}. However, the retrieval-based approach might only search targets limited to those existing in the database and be less scalable than the GAN-based generation technique, especially for the input data from the unseen/untrained type of material. In addition to the basic GAN structure, we add the residue-fusion module and the feature-matching loss to further guide the data generation between vision and touch.

\subsection{Visual-Tactile Signal Generation}
While capturing the tactile signal on the surface of an object could be difficult sometimes due to the lack of proper sensors, it is relatively easier to obtain the visual image of the object using an ordinary camera. Thus, researchers have explored the feasibility of generating tactile signals from visual images. As a preliminary attempt, Ujitoko and Ban \cite{ujitoko2018vibrotactile} apply the basic GAN structure for generating vibration-signal spectrograms from the image data or the material attributes. Liu et al. \cite{9018269} propose a CycleGAN-based framework for vibrational-signal generation based on the image data. Both of these two works utilise the LMT-108-Surface-Materials database \cite{7737070}, which includes the surface-texture images and the accelerometer data of a pen sliding on the corresponding surface from different directions, forces and velocities.

Visual-tactile data generation has also been applied to enhance the robotic capability of understanding the real world through both ``seeing'' and ``touching''.  Takahashi and Tan \cite{takahashi2019deep} propose an encoder-decoder network structure to estimate the vibrotactile properties from the surfaces' visual images. Heravi et al. \cite{heravi2020learning} introduce a learning action-conditional model to predict the acceleration/vibrational signals from the GelSight tactile images and user's actions (e.g., surface pressure and velocity). Purri and Dana \cite{purri2020teaching} develop a cross-modal adversarial framework to estimate different types of tactile properties from a set of visual images captured above a textured surface from multiple views. 

These works mainly focus on synthesizing tactile information from visual information. The other equally important direction of signal generation, from tactile to visual, is relatively less investigated. In this paper, we investigate the two-way cross-modal signal generation between the image data of a textured surface and the acceleration data of sliding a pen on such a surface, to explore the association of robotic vision and touch.

\begin{figure*}[htbp] 
\centering 
\includegraphics[width=\textwidth]{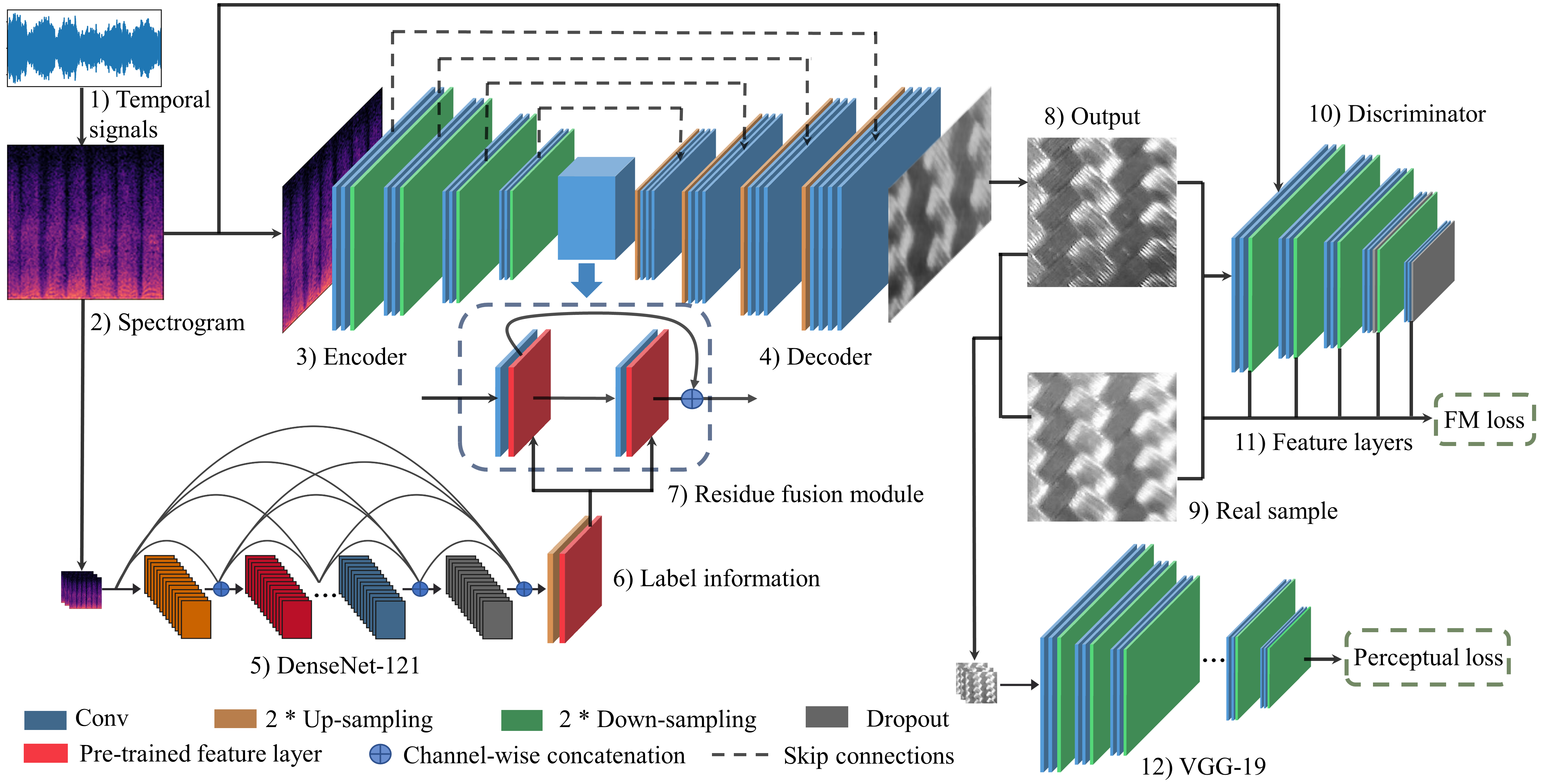} 
\caption{The overview of our proposed model. Here, we use T2V translation as our main illustration. The 1) input temporal acceleration signals are processed as 2) spectrogram data passed to 3) the encoder and 4) the decoder. We build a 7) residue-fusion (RF) module with 9 residual blocks in the latent space (The light blue cuboid). A pre-trained tactile classifier (it will be the visual classifier for V2T), 5) DenseNet-121, extracts 6) the label information and passes it to an up-sampling layer for the channel-wise concatenation with the encoder output in 7) the RF module. The decoder synthesizes 8) the output image based on the information from the latent space and passes the concatenation of the input spectrogram and the real/generated image to 10) the discriminator for the conditional adversarial training. We also extract outputs from 11) feature layers of the discriminator for feature-matching (FM) loss calculation and include 12) a pre-trained VGG-19 for calculating the perceptual loss.}
\vspace{-0.1in}
\label{Fig.2} 
\end{figure*}

\section{Methodology}
We aim to study the translation between the visual and the vibrotactile domains, which could be defined as a cross-modal data-generation problem. Inspired by the existing work on cross-modal data generation  \cite{lee2019touching, li2019connecting}, we adopted the structure of conditional GAN (cGAN) \cite{mirza2014conditional} as the base for two-way visual-tactile data generation, and enhanced the GAN structure with extra features. 

\subsection{Network Structure}
Fig. \ref{Fig.2} shows the structure of our T2V network. The architecture of V2T is similar to the T2V network structure but with the reversed input-output pair. Adopting the basic cGAN structure, our network consists of a Generator $G$ and a Discriminator $D$. We first convert the acceleration-based tactile signal to the amplitude spectrogram, which can be treated as a single-channel 2D image/matrix for the tactile domain, and use the grey-scale surface image for the visual domain. For the generator $G$, we adopt the U-net structure \cite{ronneberger2015u} as the backbone, with the skip connections between each layer \textit{i} and the layer \textit{n-i}, where \textit{n} is the total number of layers in the generator $G$. Such a structure is chosen due to its effective usage in existing image-generation research \cite{isola2017image, chen2020deepfacedrawing}. For the encoder and the decoder in $G$, we adopt a $4\times4$ kernel with the stride of 2 and the padding of 1 for each down- or up-sampling layer. In addition, we include the instance-normalization and the ReLU units in each layer of $G$. A pre-trained standard DenseNet-121 network \cite{huang2017densely} is used for extracting the label information for the residue-fusion module, which we will describe in more detail later. Our structure of the discriminator $D$ is adopted from PatchGAN \cite{isola2017image}. Each down-sampling layer in the discriminator contains a $4\times4$ convolutional kernel with the stride of 2 and the padding of 1, layer-normalization and the Leaky ReLU units.

\subsection{Residue Fusion}
Previous works on cross-modal data generation \cite{duan2019cascade, chen2017deep} show that adding domain information for strong supervision could guide the generator to synthesize the reasonable output. This method could help to solve the weak geometrical alignments between two different modalities, such as the visual and the tactile domains \cite{li2019connecting}. However, simply restricting the generated output data through label identification \cite{ujitoko2018vibrotactile} might affect the scalability of the generative model, and limit the predictive ability for new/unseen input. Inspired by recent works on high-resolution image generation \cite{chen2020deepfacedrawing, wang2018high}, we introduce the residue-fusion (RF) module in our generative model to extract more label information from the input modality. Such residual information is used as additional supervision to guide the decoder to output in the target domain. Specifically, we set up a DenseNet-121 network \cite{huang2017densely} pre-trained with ImageNet dataset \cite{imagenet_cvpr09}, and fine-tune it through transfer learning as the classifier for data samples from the input domain/modality. We remove all fully-connected layers at the standard DenseNet-121 network to extract the feature representation of label information. Then we up-sample this information of label feature and concatenate it with the feature vector from the encoder through residual blocks as the residue fusion module in the generator.

To this end, for the cross-modal paired samples $\left\{ ({x},{y}) \right\}$ where $x \in X$ and $y \in Y$ ($X$ - input and $Y$ - output can be either the visual or the tactile modality depending on the generation direction), the generator network is denoted as: $G_{\textit{X}\mapsto\textit{Y}}: \mathbb{R}^{\vert \Phi(x) \vert} \times \mathbb{R}^{\vert \Psi(x) \vert} \mapsto \mathbb{R}^{y}$, where $\Phi(x)$ is the encoded information and $\psi(x)$ represents the residual label information from the fine-tuned DenseNet-121 classifier. In our experimental settings, we adopt the residue-fusion setting for the channel-wise concatenation between the feature vectors $\Phi(x)$ and $\psi(x)$ through 9 layers of residual blocks. 

\subsection{Feature-Matching Loss}

While the traditional pixel-wise loss ($L_{1}$ or $L_{2}$ loss) could obtain considerable results for image generation \cite{isola2017image, zhu2017unpaired}, the feature-matching loss \cite{larsen2016autoencoding} also shows strong support for the same purpose (i.e. GAN-based image generation). Inspired by the previous work on image-based audio generation \cite{kumar2019melgan}, we include the feature-matching (FM) loss into the model-training process to extract the feature outputs from multiple layers of the discriminator and match these feature representations from the real and the generated outputs. Specifically, the feature-matching loss ${L}_{fm}$ in our model is calculated as:

\begin{equation}
\textit{L}_{fm} = \mathbb{E}_{y \sim p_{(y)},\tilde{y} \sim p_{(\tilde{y})}} \sum_{i=1}^{T} \frac{1}{N_{i}} [\|D^{(i)}(y)-D^{(i)}(\tilde{y})\|_{1}],
\label{1}
\end{equation}

In this equation, $y$ and $\tilde{y}$ indicate the real and the generated samples while $p{(y)}$ and $p{(\tilde{y})}$ represent the distribution of real and generated data, respectively; we denote $D^{(i)}$ as the features in the $i$-th layer of the discriminator $D$, $T$ is the total number of layers in $D$, and $N_{i}$ is the number of elements in each layer $D^{(i)}$.

\subsection{Perceptual Loss}
To further extract the feature of both visual and tactile data, we also include the perceptual loss \cite{johnson2016perceptual} that is calculated from the outputs of multiple layers through VGG-19 pre-trained by ImageNet \cite{imagenet_cvpr09}. Specifically, we optimize the $L_{2}$ distance of the outputs from the feature layers (denoted as $F_{vgg}$) between the generated and the real data. Similarly, $M$ is the elements' number in each feature layer, and the perceptual loss ${L}_{p}$ is defined as:

\begin{equation}
\textit{L}_{p} = \mathbb{E}_{y \sim p_{(y)},\tilde{y} \sim p_{(\tilde{y})}} \frac{1}{M}[\|F_{vgg}(y)-F_{vgg}(\tilde{y})\|_{2}],
\label{2}
\end{equation}

\subsection{Objective Function}
In the original GAN, the objective function for the generator may cause gradient vanishing while the network being trained without carefully adjusting the hyper-parameters \cite{arjovsky2017wasserstein}. To increase the network robustness and prevent model collapsing, we adopt the WGAN-GP loss \cite{gulrajani2017improved} as $\textit{L}_{adv}$, to optimize the Wasserstein distance between the ground truth and the generated data. Hence, combining the feature-matching loss and the perceptual loss, our objective function of the proposed method is shown below:

\begin{equation}
\mathop{\arg\min \limits_{G} \max \limits_{D} \quad \textit{L}_{adv} + \alpha \textit{L}_{fm}} + \beta \textit{L}_{p} 
\label{3}
\end{equation}

This formulation includes the adversarial loss $\textit{L}_{adv}$, the feature matching loss $\textit{L}_{fm}$ and the perceptual loss $\textit{L}_{p}$. We set the parameters $\alpha$ as 10 and $\beta$ as 1 in the Eq. \ref{3} for T2V data generation, which is similar to the previous image-to-image generation work with FM loss \cite{wang2018high}, and empirically determined $\alpha$ = 100, $\beta$ = 10 for V2T cross-modal data generation. The larger values of V2T parameters is because the spectrogram data usually contains fewer features than visual images, which means there are more values close to 0 in the 2D matrix of the spectrogram data, so the loss of the feature outputs for training the V2T model would yield lower values than the T2V model, needing larger values of the parameters to accelerate the model convergence.

\section{Data Preparation}
The LMT-108 Surface-Materials database \cite{7737070} provides both the visual images of different surfaces and the acceleration signals induced by the pen-sliding movements of robotic arms and human hands on the corresponding surfaces. It has been used in the previous works of GAN-based acceleration-signals generation \cite{ujitoko2018vibrotactile, 9018269}. This database contains 108 types of surface materials being grouped into 9 categories. Each category of materials includes 20 sets of RGB surface images and acceleration-based tactile signals. Each set of tactile data of a material type contains the time-series acceleration signals in three directions (i.e., X, Y, and Z). Referring to the previous works on visual-to-tactile signal generation \cite{ujitoko2018vibrotactile}, we randomly select one type of material from each category, thus a totally 9-class/type subset from the overall 108 types of materials, as our experimental database. This set includes Squared Aluminum Mesh (M1), Marble (M2), Acrylic Glass (M3), Compressed Wood (M4), Fine Rubber (M5), Carpet (M6), Fine Foam (M7), Carbon Foil (M8), and Leather (M9).  

\textbf{Tactile data.} 
Similar to the previous work on image-to-tactile generation \cite{ujitoko2018vibrotactile}, we focus on the vibrotactile signal on the Z-axis, which demonstrates the most obvious vibration during the pen-sliding movements. As GAN has been widely adopted for image generation, we first convert the time-series tactile signals to the format of amplitude spectrogram, which could be represented as a 2D image/matrix. Each original tactile signal lasts 4.8-s with a sample rate of 10kHz captured by the accelerometer in a pen-based device. Thus, the signal could be affected by the starting point and the initial pressing force. To reduce such variation, we remove the signal in the first second and convert the remaining 3.8-s signals into the $257\times297$ spectrogram using Short-Time Fourier Transform (STFT) algorithm with a 512-Hamming window and a 128-hop size. Lastly, we randomly crop each spectrogram along the time axis to the size of $256\times256$ corresponding to 3.24-s acceleration signals and scale it logarithmically as our tactile data set.

\textbf{Visual data.}
The original LMT-108 Surface-Materials database includes visual images with or without flash condition in the collection process. Following the previous work on cross-modal learning for material perception with the same database \cite{2019Cross}, we focus on the images captured without the flashlight. For data augmentation, we randomly flip each image horizontally and vertically, adjust the parameters of contrast and brightness for each image, and crop each image to the size of $256\times256$ from the original $640\times480$ texture image. The similar data-augmentation methods for visual images (e.g., flipping, brightness and contrast adjustment) cannot be applied to tactile data, as these processes may affect the temporal characteristics and amplitude strength of acceleration signals \cite{9018269}. We then convert the RGB images into 1-channel images as the touch sense on a surface may not strongly depend on surface colours.

\textbf{Weakly paired data.}
The backbone of our proposed model is initially designed for image-to-image generation. However, in the case of cross-modal visual-tactile data generation, the data-collection procedures in the two modalities are independent \cite{7737070}. Therefore, it is not trivial to construct the exact one-to-one correspondence between the visual and the tactile data. To this end, it is proposed to adopt the weakly data-pairing strategy \cite{lampert2010weakly, liu2017weakly} for cross-modal visual-tactile data generation. In our case, to create the weakly pairing mechanism between the visual and the tactile domains, we repeat the data-pre-processing and -augmentation procedures 100 times for all the original visual and tactile data within each selected material type. Thus, we obtain 20 sets of visual-tactile data-pairs, with each set of 100 augmented and randomly-paired data as our weakly paired data, for each type of material. One may argue that the weakly paired data between the tactile spectrograms and the visual images may discard the phase information of the pen-sliding motion on the textured surface, leading to the potentially similar tactile representations for two different materials. While this could be true, existing research \cite{zheng2019cross} actually shows that the weakly paired data could be more effective for cross-modal data generation under a weakly-controlled data-collection process adopted by the LMT-108 database. Furthermore, existing psychophysical research shows that the phase shift of the on-surface motion places a less important effect on humans' surface-texture sensation, compared to the general time-frequency feature \cite{heravi2020learning, cholewiak2009frequency}. As a result, we acquire a total of $20\times100\times9=18000$ weakly paired visual-tactile data (i.e., images and spectrograms) and normalize them to the range of -1 to 1. For the visual images and the spectrograms of each selected material, we randomly split the data with the ratio of 8 : 1 : 1 (training : validation : testing).

\begin{figure}[t]
\centering
\begin{minipage}{\linewidth}
\centering
\includegraphics[width=\columnwidth]{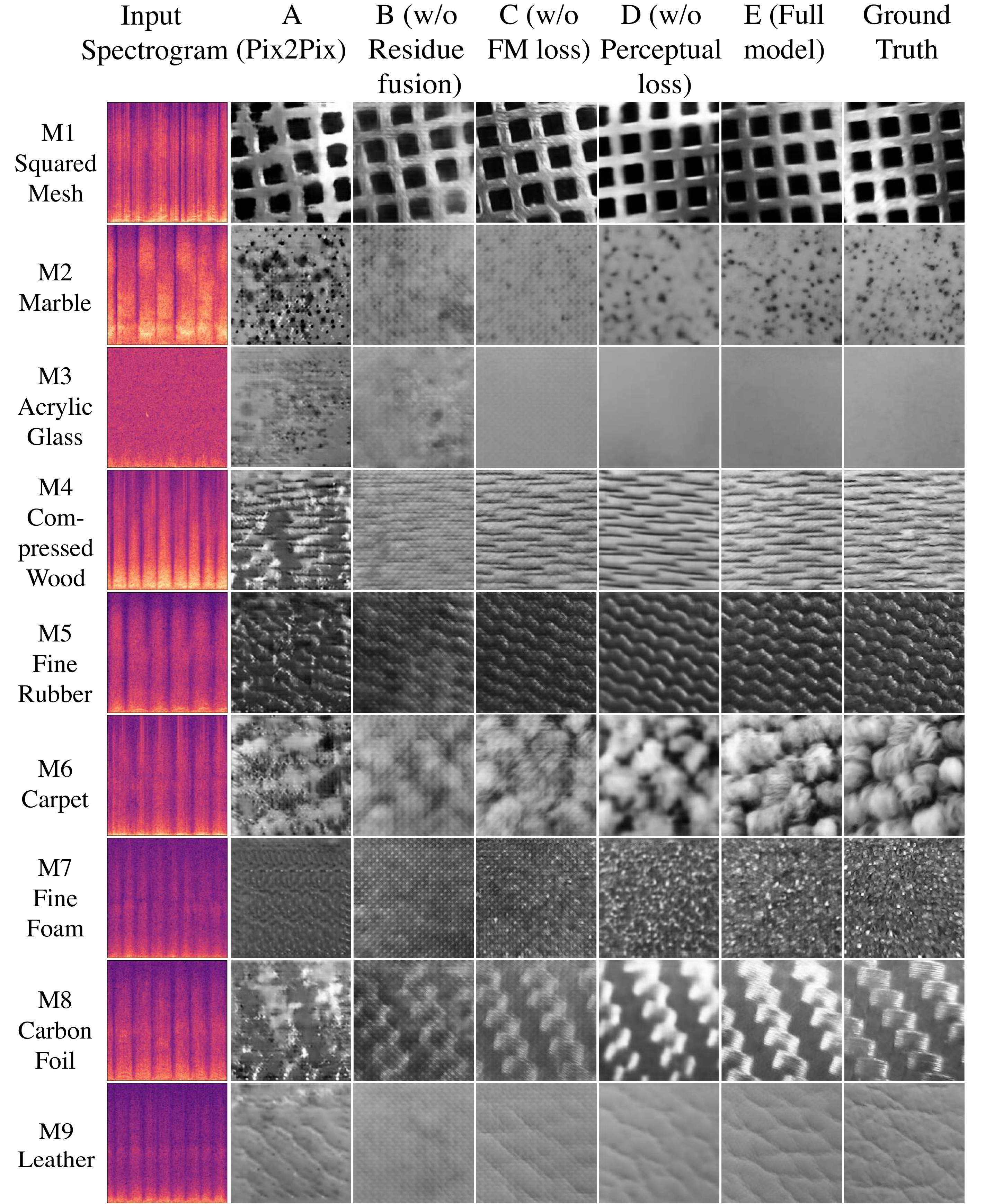}
\end{minipage}%

\caption{Examples of T2V cross-modal prediction results. The first column represents the input spectrograms (M1-M9 totally 9 categories material). The second to fifth columns show all generation results from different T2V Models (from Model A to Model E), based on the corresponding input spectrograms. The last column shows the ground truth targets in cross-modal predictions.} 
\vspace{-0.12in}
\label{Fig.4} 
\end{figure}

\begin{figure}[t]
\centering
\begin{minipage}{\linewidth}
\centering
\includegraphics[width=\columnwidth]{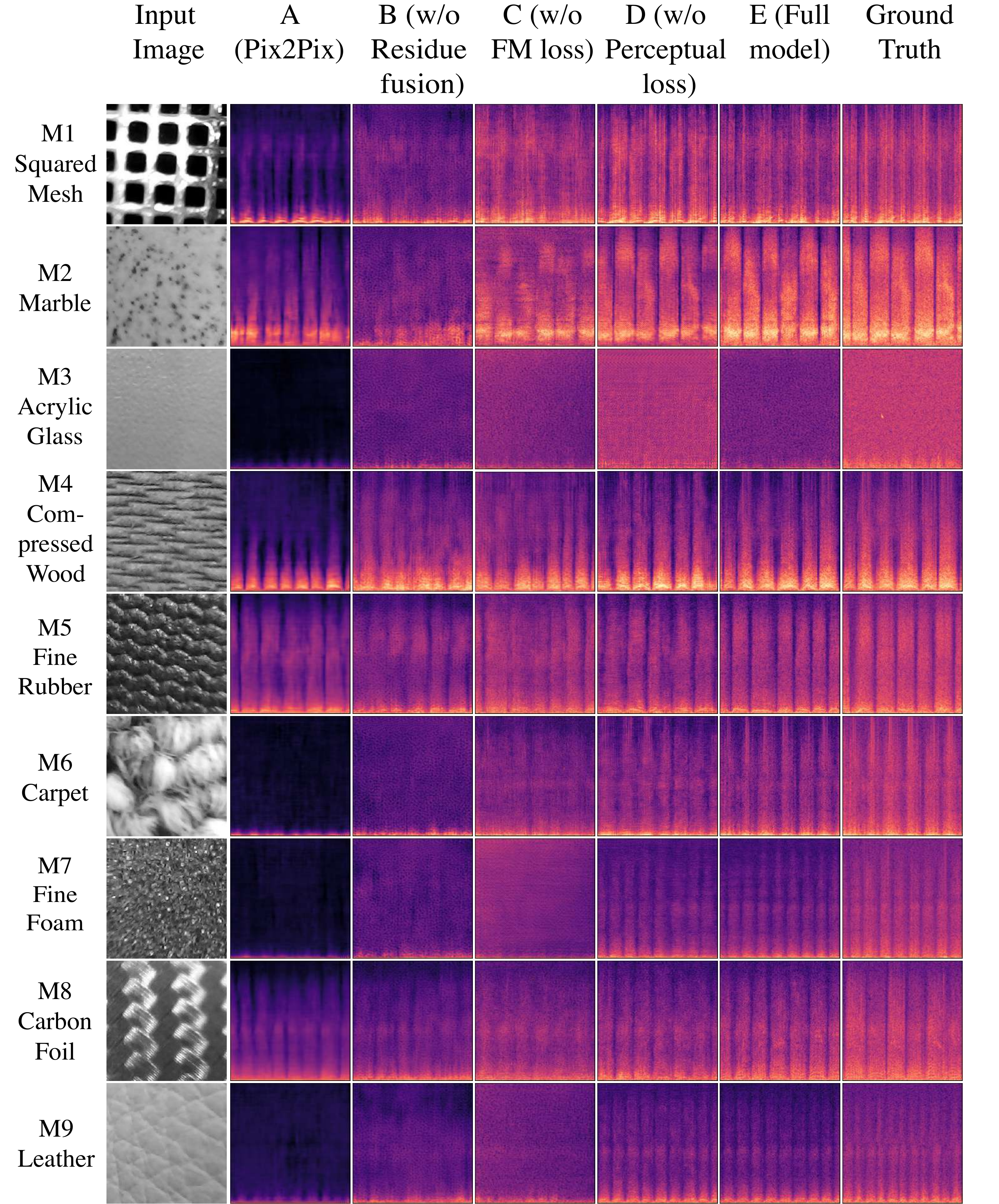}
\end{minipage}%
\caption{Examples of V2T cross-modal data generation results with the similar layout of Fig. \ref{Fig.4}.}
\vspace{-0.12in}
\label{Fig.5} 
\end{figure}

\section{Experiments}
\subsection{Experimental Settings}
\textbf{Baseline Selection.}
To evaluate the capabilities of our framework, we compare the generative results from our model, and the Pix2Pix model \cite{isola2017image} which was used in the previous work on robotic cross-modal vision-to-touch generation \cite{lee2019touching}. Some other visual-to-tactile generation models \cite{ujitoko2018vibrotactile, 9018269} only support the single-direction data generation (i.e. from vision to touch) and lower resolutions (e.g., $128\times128$), which are not fair to be used for our comparison. We also conduct the ablation study to verify the effectiveness of our model's key components (i.e., the residue-fusion module, the feature-matching loss, and the perceptual loss). 

\textbf{Evaluation Metrics.}
Following the evaluation method adopted by Lee et al. \cite{lee2019touching}, we first evaluate our model by classifying the generated visual and tactile data with the pre-trained visual and tactile classifiers (DenseNet-121-based), respectively. As the second evaluation metric, we compute the Frechet Inception Distance (FID) \cite{heusel2017gans} between the ground-truth data and the generated data. The FID evaluation metric is widely used for evaluating GAN performance \cite{borji2019pros}.

\textbf{Training Process.}
We first train two DenseNet-121 networks \cite{huang2017densely} to classify the tactile spectrograms and the visual images separately. These two classification networks are used for residual fusion (Fig. \ref{Fig.2} part 4) and later model evaluation. To avoid overfitting, we perform the data augmentation on the tactile signals based on the time and frequency masking method \cite{park2019specaugment} and add random Gaussian noise into the visual images during the classifiers training stage. The values of elements in both images and spectrograms are normalized in the range from $0$ to $1$ for classifier training. Both classifiers are trained with Adam optimization and 1e-4 learning rates. We achieve 99.32$\%$ classification accuracy for the visual images and 96.22$\%$ classification accuracy for the tactile spectrograms on the testing data sets. We then freeze the parameters of these two classifiers and embed them into the V2T and the T2V generative models, respectively (i.e., tactile classifier for T2V and visual classifier for V2T), for the setup of the residue fusion module. For the Pix2Pix model \cite{isola2017image}, we follow the similar structural and training settings in Lee et al.'s work \cite{lee2019touching}. 

All training and testing experiments are implemented with TensorFlow 2.1.0 framework on an Nvidia Geforce GTX 2080Ti GPU with batch size = 8. We set the learning rates of generator and discriminator as 2e-4, with the Adam optimizer ($\beta_1$ = 0.9 and $\beta_2$ = 0.999). All model weights are initialized with Xavier normal initializer \cite{glorot2010understanding}.

\subsection{Comparison Study}
\textbf{Study Settings.}
In our comparison study, we implement five different models among our augmented database. Model A: Pix2Pix model \cite{isola2017image}, which is used as the baseline for the comparison. Model B removes the residue-fusion module of the generator. Model C and D remove the feature matching loss ${L}_{fm}$ and the perceptual loss ${L}_{p}$, respectively. Model E is our full model with the residue-fusion module, the feature-matching loss, and the perceptual loss. Model B, C and D are implemented for ablation study to study the effectiveness of the key components in our full model.

\textbf{Baseline Comparison.}
Fig. \ref{Fig.4} and Fig. \ref{Fig.5} visually illustrate all generated results of T2V and V2T generation, respectively. Compared to our baseline Model A, the full Model E leads to improved visual quality both on texture images and amplitude spectrograms. The classification accuracy and FID scores (Table \ref{table1}) echo with the visual comparison. Specifically, the pre-trained visual classifier achieves an overall accuracy of 94.61\% with the visual images generated by the Model E, leading to the largest improvement over the Model-A-generated images (94.61\% vs 53.78\%). The visual data generated by Model E achieves the FID score of 110.11, with 0.55 decrements comparing to the baseline model (FID: 245.19). For the tactile data generation (i.e. V2T), the Model-E generated data yields an overall classification accuracy of 83.78\% (baseline: 40.90\%), and the lowest FID score of 48.40 (baseline: 165.60).

\begin{table}[]
\caption{\label{table1}The Evaluation Metrics (EM) results: Classification Accuracy (CA) and FID values from different models (Model A-E).}
\resizebox{\columnwidth}{!}{
\begin{tabular}{|c|c|c|c|c|c|c|}
\hline
EM &  Tasks  & A      & B      & C      & D      & E      \\ \hline
\multirow{2}{*}{CA} & T2V & 53.78\% & 47.89\% & 69.67\% & 81.78\% & \textbf{94.61\%} \\ \cline{2-7} 
                     & V2T & 40.89\% & 41.89\% & 38.22\% & 71.44\% & \textbf{83.78\%}\\ \hline
\multirow{2}{*}{FID} & T2V & 245.19 & 251.89 & 194.07 & 182.81 & \textbf{110.11} \\ \cline{2-7} 
                     & V2T  & 165.60 & 104.43 & 106.65 & 95.28 & \textbf{48.40} \\ \hline
\end{tabular}
}
\vspace{-0.15in}
\end{table}

\textbf{Ablation Study.} For the T2V generation, if we remove the residue-fusion module (Model B), the FM loss (Model C) and the perceptual loss (Model D), the classification accuracy are 47.89\%, 69.67\%, and 81.78\%, while the FID scores are 251.89, 194.07, and 182.81, respectively. To this end, the residue-fusion module plays an essential role in the T2V generation by improving classification accuracy with a ratio of 0.97 and 1.29 in FID compared to Model B missing the RF module. This result further indicates the effectiveness of adding the residual label information for supervising reasonable output. 

Similarly, the RF module also shows a strong influence on V2T cross-modal data generation, which leads to an approximately doubled improvement in classification accuracy and a decrement ratio of 0.54 in FID (Model E vs Model B). Unlike T2V, where the RF module shows the most dominant effect, the FM loss makes the most influential effect on tactile data generation in our ablation study. The generated spectrograms only acquire the classification accuracy of 38.22\% and 106.65 in FID without the FM loss. This result suggests that the FM loss outperforms the traditional pixel-wise loss ($L_{1}$ or $L_{2}$ loss) on spectrogram generation for tactile signals.

\subsection{Data Generation for Different Materials}

\begin{figure}[tb]
\centering
\subfigure[The confusion matrix in T2V Generation]{
\begin{minipage}{\linewidth}
\centering
\includegraphics[width=0.97\columnwidth]{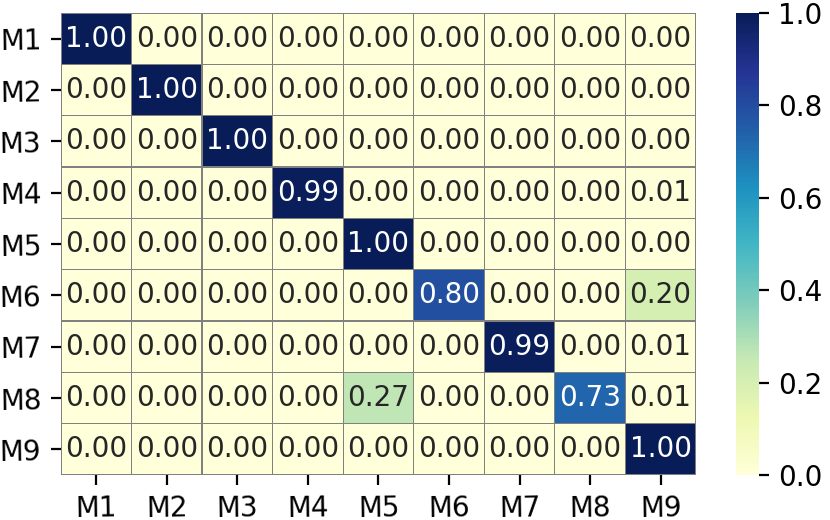}
\vspace{0.05in} 
\end{minipage}%
}%
\vspace{-0.1in} 
\subfigure[The confusion matrix in V2T Generation]{
\begin{minipage}{\linewidth}
\centering
\includegraphics[width=0.97\columnwidth]{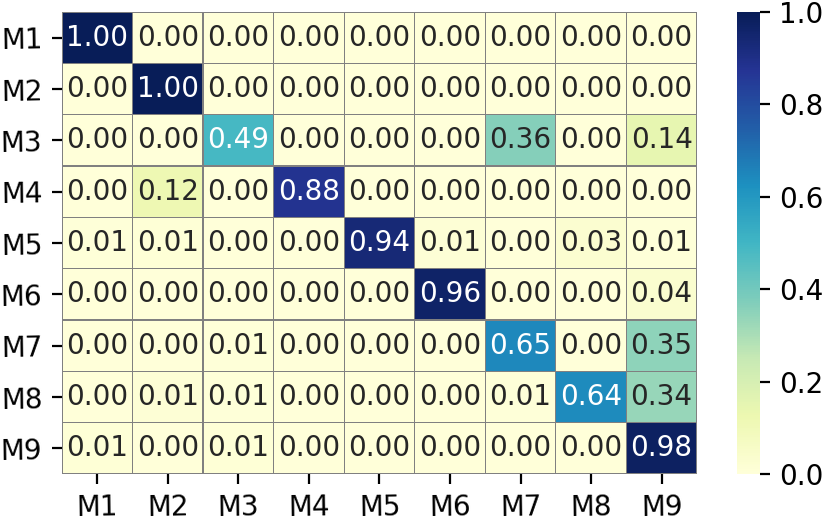}
\vspace{0.02in} 
\end{minipage}%
}%
\vspace{-0.1in} 
\caption{The confusion matrix of generated data from our full models both in T2V (a) and V2T (b) generation. All rows represent the ground-truth labels and columns are the predictions by the classifiers.} 
\vspace{-0.1in}
\label{Fig.6} 
\end{figure}

\textbf{Tactile $\rightarrow$ Visual}: Fig. \ref{Fig.4} shows that the full-model-generated visual data of different categories of materials tend to be realistic and close to the ground truth images. The average classification accuracy upon the generated visual images is 94.61\%, close to the testing accuracy with the ground-truth data (99.32\%). On the other hand, the confusion matrix in Fig. \ref{Fig.6} (a) reveals that two types of materials obtain the lower accuracy: M6-Carpet (79.50\%) and M8-Carbon Foil (73.00\%). The possible reason is that the carpet (M6) visual patterns appear more irregular and diverse, which may confuse the pre-trained visual classifier. The Carbon Foil (M8) generated data yields the worst performance on the visual-image-based material classification. This might be due to the lighting condition during the data collection, as the Carbon Foil (M8) is reflective and the images were captured under different lighting conditions \cite{7737070}. This finding suggests that the diversity of lighting conditions should be considered while collecting cross-modal data and designing robots with visual-tactile fusion capability.

\textbf{Visual $\rightarrow$ Tactile}: Fig. \ref{Fig.6} (b) depicts the confusion matrix of the generated tactile signals using our full V2T model. The classification accuracy of Acrylic Glass (M3), Fine Foam (M7) and Carbon Foil (M8) is 49.00\%, 64.50\% and 64.00\%, respectively. The lowest classification accuracy yielded by M3 could be due to the lack of significant features both in the visual and the tactile signals. In addition, the tactile signals of M3 might be affected by noise (possibly from the stain on the material surface or the hand movements during the pen-sliding process). Thus, it is difficult to train the generative model for tactile data from images with the missing feature. The low classification accuracy of M7 and M8 could be due to their tactile similarity to the other materials. For instance, the ``Ground Truth'' column of Fig. \ref{Fig.5} shows that the amplitude spectrogram of M7 appears to be visually similar to that of Leather (M9). The confusion matrix in Fig. \ref{Fig.6} (b) also shows that there are 35.40\% of the generated Fine Foam (M7) tactile data samples classified as the Leather (M9). We further perform the Dynamic Time Warping (DTW) algorithm upon the tactile signals of 9 categories of material. The warping distance between M7 \& M9 and M8 \& M9 are 5.32 and 5.46, respectively, which are the lowest among all materials (averagely 17.93), indicating the similarity between M7 \& M9 and M8 \& M9, which might confuse the classifier. We further examine the intra-class variance \cite{pilarczyk2019intra} for the tactile data of each material type. The results show that M9 obtains a larger average variance (15.22) than both M7 (3.99) and M8 (3.41), making the tactile classifier biased to the class with a larger data variation (i.e. M9) \cite{holte1989concept}.

\section{Discussion}
The above experimental results show that our generative model outperforms the baseline model and other ablation-study models in both T2V and V2T generation. Our method aims to support robotic operational tasks through complementary modalities. For example, a robot could imagine the tactile characteristics of an object to adopt a better grasping strategy. The robotic device could also generate the visual information based on the tactile perception to improve the recognition performance during the low-light condition as the vision-based recognition usually yields more considerable performance. Based on our configuration, both V2T and T2V generations take about 0.04-s, and 0.02-s for the visual/tactile recognition through the classifiers, for each sample. It takes 3.24-s for acquiring sufficient time-series tactile data and converting it to the spectrogram for T2V generation while cameras can capture the images in real-time for V2T. Such time duration is comparable to the human's exploration process for surface identification \cite{kappers2011human}.

In the experiment, we can see that the Pix2Pix model \cite{isola2017image} yields the worst performance on cross-modal visual-tactile data generation. The possible reason could be that the Pix2Pix model usually requires geometrical alignments between the input and the output domains, such as the RGB and the GelSight-based images \cite{lee2019touching}. Such alignments could be obscure in the weakly paired visual-tactile data. The previous works also show that the Pix2Pix model may lead to blurry or distorted results on visual data generation (e.g., the fabric images \cite{lee2019touching} and the visual image of robot position \cite{li2019connecting}).

We also notice that the performance of the T2V generation outperforms the V2T generation for the classification evaluation, which could be due to the backbone framework is originally designed for visual image generation. Although the spectrograms could be treated as 2D matrices during the training process, they are essentially different from visual images. While an image is usually treated as a 2D matrix with each element ranging from 0 to 255, the range of the spectrogram elements could be (0, +$\infty$). The normalization process may push the normalized values too close to the range boundary, so the average pixel-wise loss may not fully reflect the actual data distribution. We will investigate different model designs specifically for the spectrogram data in future work to improve the V2T cross-modal generation. 

\begin{figure}[t]\centering                                         \subfigure[Less successful cases]{              
\begin{minipage}{0.5\columnwidth}
\centering                                             \includegraphics[scale=0.37]{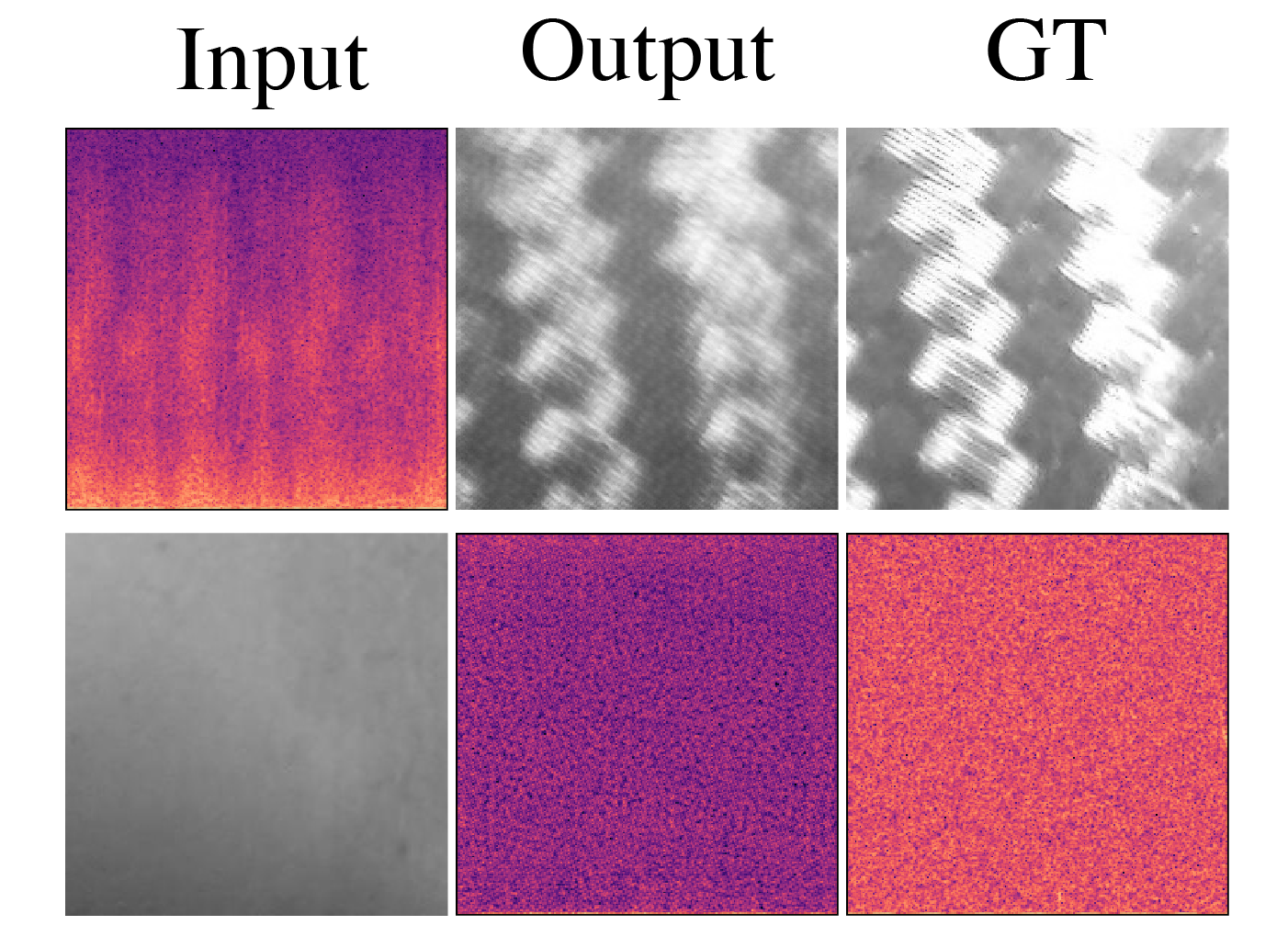}       \end{minipage}}\subfigure[New materials prediction]{                \begin{minipage}{0.5\linewidth}
\centering                                           \includegraphics[scale=0.37]{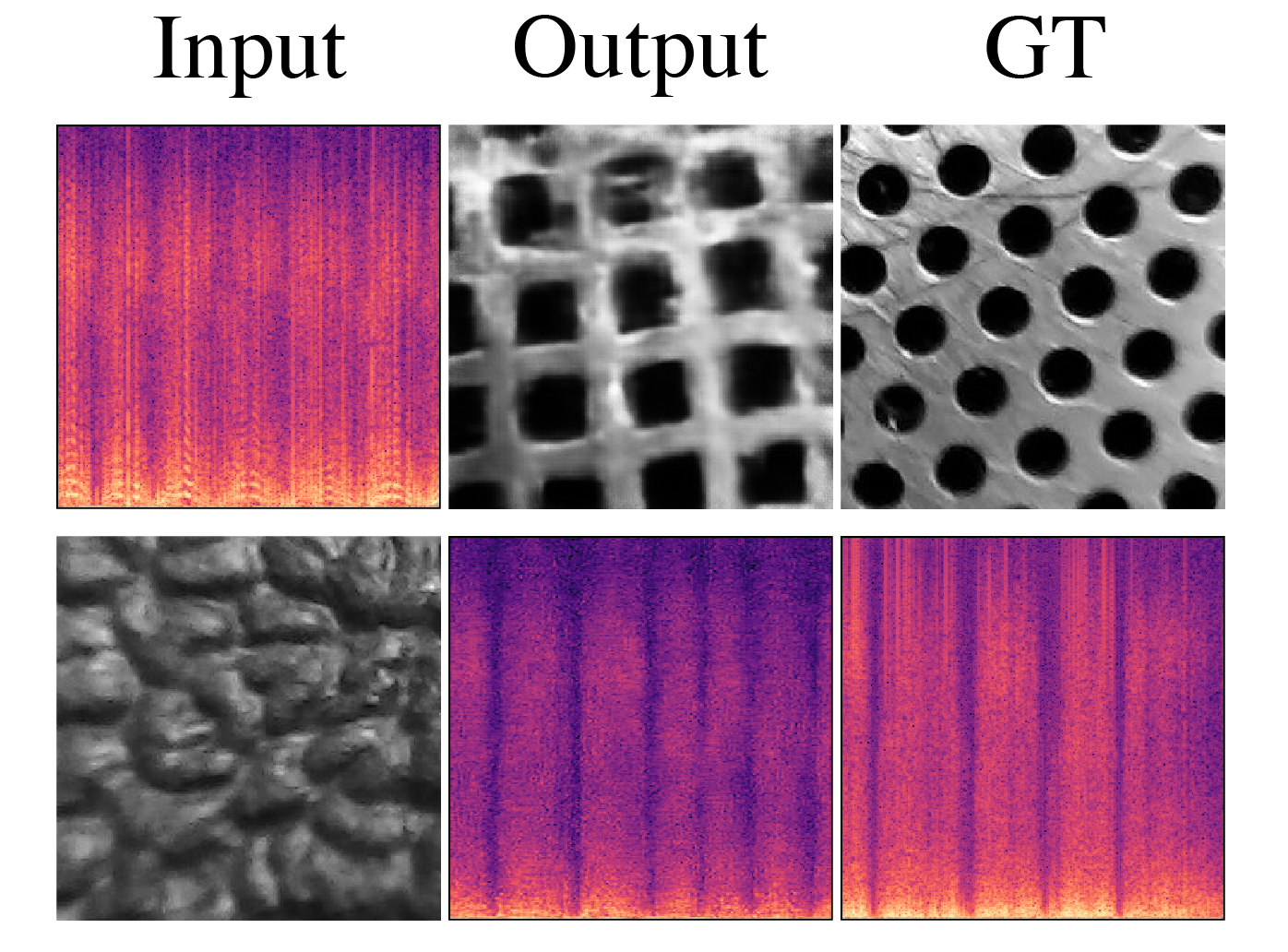}  
\end{minipage}}
\vspace{-0.1in} 
\caption{Examples of (a) less successful cases and (b) generation results for new materials. 
}    
\vspace{-0.22in}
\label{fig:7}                                   
\end{figure}

In addition, we observe a few less successful cases in T2V and V2T generation. Fig. \ref{fig:7} (a) shows two less successful examples which might be due to the harsh lighting (first row) and the noisy vibrotactile signals (second row) in the ground-truth (GT) data. This could be due to the uncontrolled data-collection procedure. For instance, different ambient illumination or human-hand motion may be confounding factors to the visual and tactile features. The influence of such factors could be reduced with a more extensive and comprehensive database. Besides, the original LMT-108 database did not label the specific configurations of pen-sliding (e.g., direction, force, and velocity) for each data entry, limiting further evaluation under different scanning conditions.

Considering the anisotropic characteristic of material-surface, one alternative data-arrangement scheme is to stack temporal acceleration signals collected from different directions/velocities/forces into a 2D matrix as our tactile data instead of spectrograms. Such tactile representations may allow generating various acceleration signals corresponding to different conditions under the same input texture image. However, it requires much more tactile data for training the generative model to match the resolution of the visual image for two-way generation using the same framework. Therefore, we plan to collect a large-scale visual-tactile database with different controlled sampling conditions (e.g., different illumination settings during the image-capturing process and different motion parameters during the tactile data collection) for the next step of model training and testing.

Lastly, we will further explore the generation of unseen visual or tactile data using our model. For intelligent robots, not only ``imagine'' texture information from what they ``see'', but also may ``create'' new texture based on their ``learned knowledge''. We test some materials that are not included in our 9-category data set, such as Circle Mesh and Profiled Rubber. The generated results are shown in Fig. \ref{fig:7} (b) (first row: Circle Mesh; second row: Profiled Rubber). The results show some similar features (e.g., similar meshed pattern or texture surface characteristics) to some extends comparing to ground-truth samples. In future work, we will investigate the influence of the latent space of our pre-trained generator for generating cross-modal results based on unseen types of input.

\section{Conclusions}
In this paper, we present a residue-fusion (RF) GAN trained with additional feature-matching (FM) and perceptual losses for cross-modal visual-tactile data generation. We validate our model upon the data of 9 types of materials selected from the LMT-108 Surface-Materials database for cross-modal V2T and T2V data generation. The results show our model outperforms the baseline model with the considerable recognition performance of the visual domain (94.61\%) and the tactile domain (83.78\%). The ablation study also reveals the effectiveness of the RF module, the FM and the perceptual losses. Our approach could be potentially applied in various robotic operational tasks, such as object recognition in low-light conditions and light-weight object grasping.


\normalem
\bibliographystyle{IEEEtran}
\bibliography{IEEEabrv,IEEEexample}
\end{document}